\documentclass[10pt,twocolumn,letterpaper]{article}

\usepackage{cvpr}              % To produce the CAMERA-READY version
\usepackage{amsfonts} %%%%New added package, check if no problem??
\usepackage{amsmath} %%%%New added package, check if no problem??
\usepackage{booktabs} %%%%New added package, check if no problem??
\usepackage{multirow} %%%%New added package, check if no problem??
\usepackage{comment} %%%%New added package, check if no problem??
\usepackage{subcaption} %%%%New added package, check if no problem??
\definecolor{cvprblue}{rgb}{0.21,0.49,0.74}
\usepackage[pagebackref,breaklinks,colorlinks,allcolors=cvprblue]{hyperref}

\def\paperID{17964} % *** Enter the Paper ID here
\def\confName{CVPR}
\def\confYear{2025}

\title{CLoCKDistill: Consistent Location-and-Context-aware Knowledge Distillation for DETRs}

\author{
Qizhen Lan and Qing Tian \thanks{Corresponding author. This work was supported by the National Science Foundation (NSF) under Award No. 2153404 and No. 2412285.}\\
University of Alabama at Birmingham \\
{\tt\small \{qlan, qtian\}@uab.edu}
}

\begin{document}
\maketitle

\begin{abstract}
Object detection has advanced significantly with Detection Transformers (DETRs). However, these models are computationally demanding, posing challenges for deployment in resource-constrained environments (e.g., self-driving cars). Knowledge distillation (KD) is an effective compression method widely applied to CNN detectors, but its application to DETR models has been limited. Most KD methods for DETRs fail to distill transformer-specific global context. Also, they blindly believe in the teacher model, which can sometimes be misleading.
To bridge the gaps, this paper proposes Consistent Location-and-Context-aware Knowledge Distillation (CLoCKDistill) for DETR detectors, which includes both feature distillation and logit distillation components.
For feature distillation, instead of distilling backbone features like existing KD methods, we distill the transformer encoder output (i.e., memory) that contains valuable global context and long-range dependencies. Also, we enrich this memory with object location details during feature distillation so that the student model can prioritize relevant regions while effectively capturing the global context. 
To facilitate logit distillation, we create target-aware queries based on the ground truth, allowing both the student and teacher decoders to attend to consistent and accurate parts of encoder memory. Experiments on the KITTI and COCO datasets show our CLoCKDistill method's efficacy across various DETRs, e.g., single-scale DAB-DETR, multi-scale deformable DETR, and denoising-based DINO. Our method boosts student detector performance by 2.2\% to 6.4\%.
\end{abstract}    
\section{Introduction}
\label{sec:intro}
Early deep learning detection approaches predominantly rely on convolutional neural networks (CNNs) to process regional features, involving extensive tuning of hand-crafted components such as anchor sizes and aspect ratios \cite{ren2015faster, lin2017focal}. Inspired by the success of transformers in NLP and visual classification tasks, DEtection TRansformer (DETR) \cite{carion2020end} has been introduced for visual detection. DETR treats object detection as a set prediction problem, eliminating the need for hand-crafted anchor design \cite{ren2015faster} and non-maximum suppression (NMS) \cite{hosang2017learning}. Despite achieving state-of-the-art performance, transformer-based detectors \cite{carion2020end, zhu2020deformable} suffer from high computational costs, making them challenging to deploy in real-time applications (e.g., autonomous driving perception).
Knowledge distillation (KD) \cite{hinton2015distilling}, a technique where a smaller model is trained to mimic a larger model's behavior, has shown to be an effective compression method for convolution-based detectors \cite{zhang2020improve, li2017mimicking, wang2019distilling, dai2021general, chen2017learning, dai2021general, zheng2022localization}. However, the exploration of knowledge distillation in DETRs is still in its early stages, with several critical issues remaining.

\begin{figure}[t]
\begin{center}
    \begin{subfigure}[b]{0.31\linewidth}
    \includegraphics[width=\textwidth]{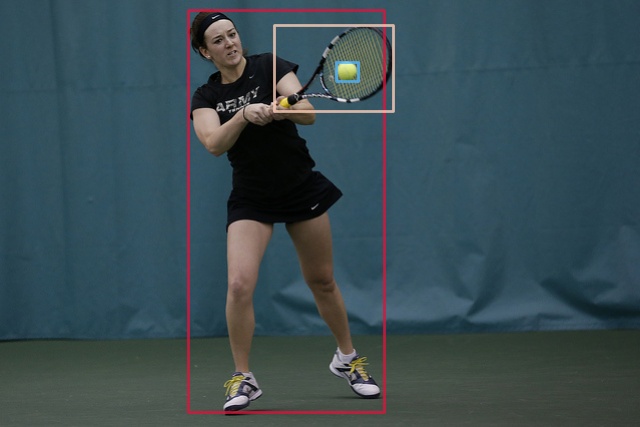}
    \end{subfigure}
    \begin{subfigure}[b]{0.31\linewidth}
    \includegraphics[width=\textwidth]{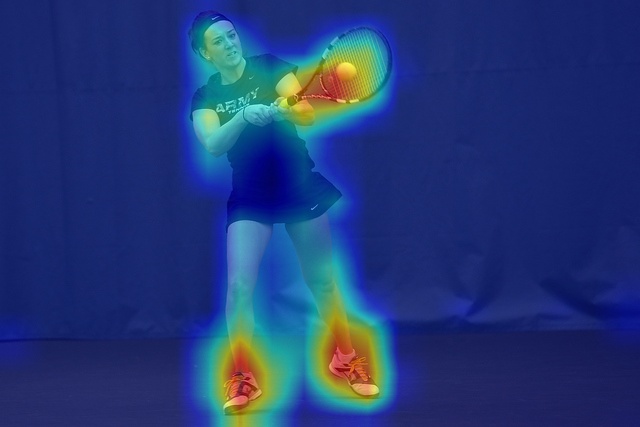}
    \end{subfigure}
    \begin{subfigure}[b]{0.31\linewidth}
    \includegraphics[width=\textwidth]{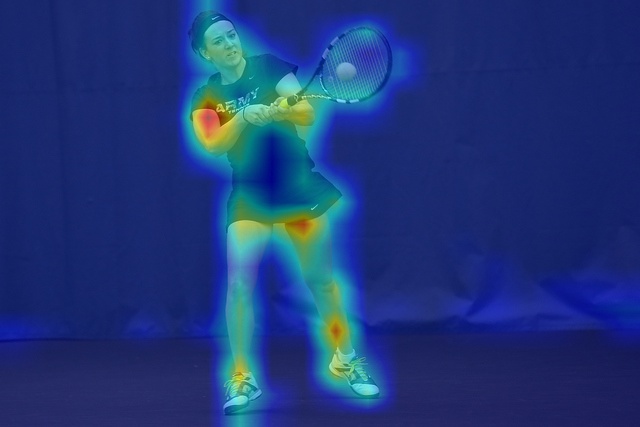}
    \end{subfigure}

    \begin{subfigure}[b]{0.31\linewidth}
        \includegraphics[width=\textwidth]{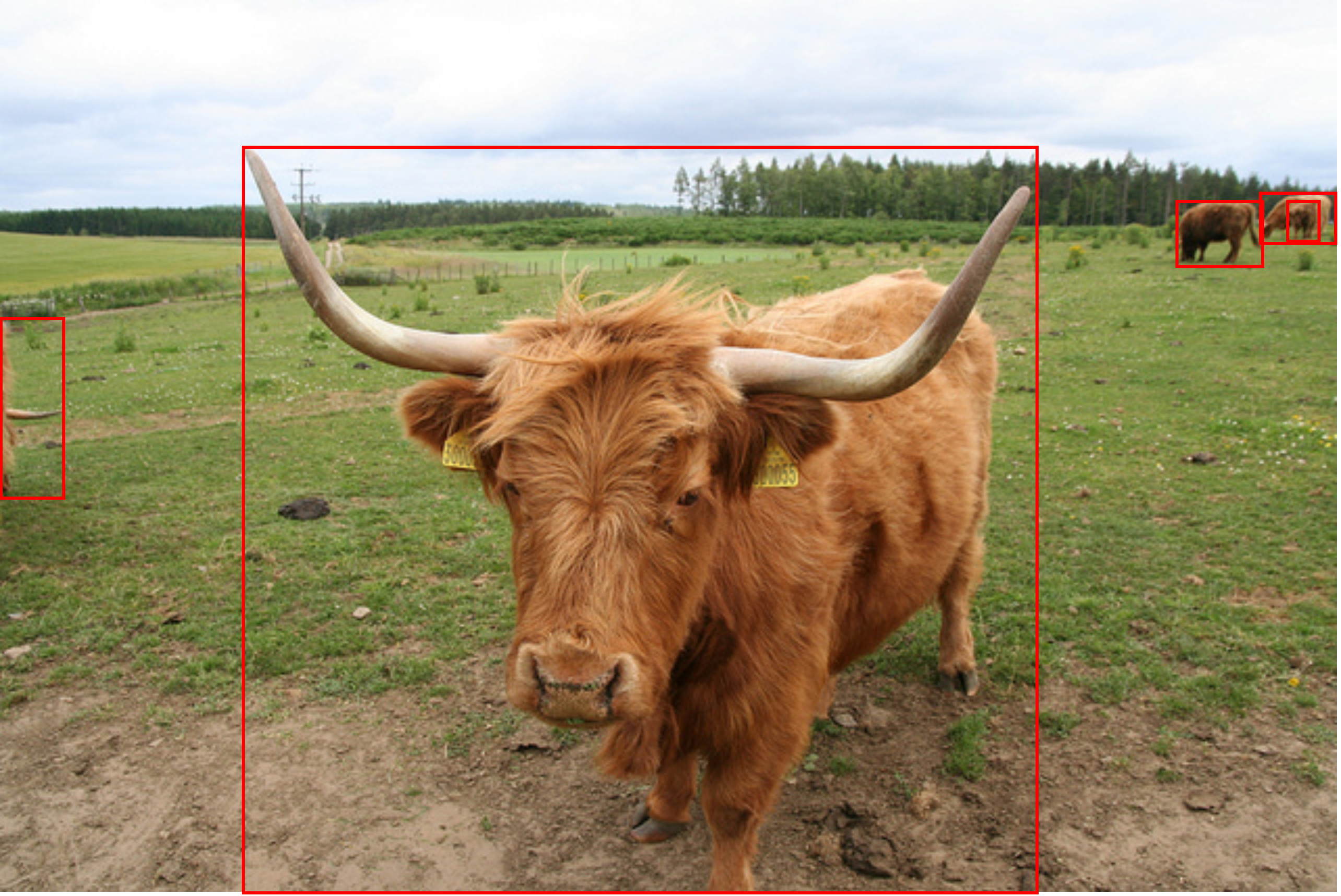}
        \caption{Image w/ bboxs}\label{fig:ori}
    \end{subfigure}
    \begin{subfigure}[b]{0.31\linewidth}
        \includegraphics[width=\textwidth]{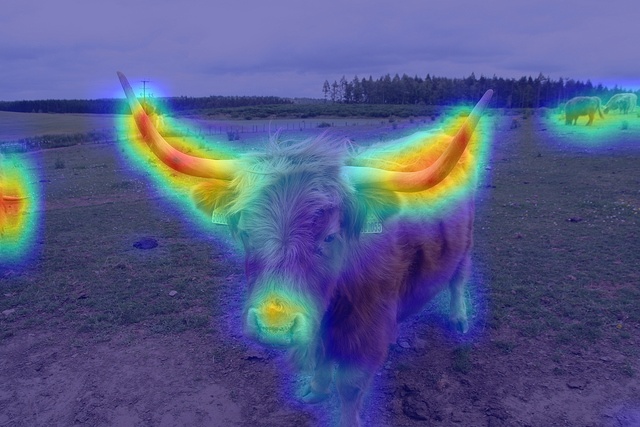}
        \caption{CNN-based}\label{fig:atss}
    \end{subfigure}
    \begin{subfigure}[b]{0.31\linewidth}
        \includegraphics[width=\textwidth]{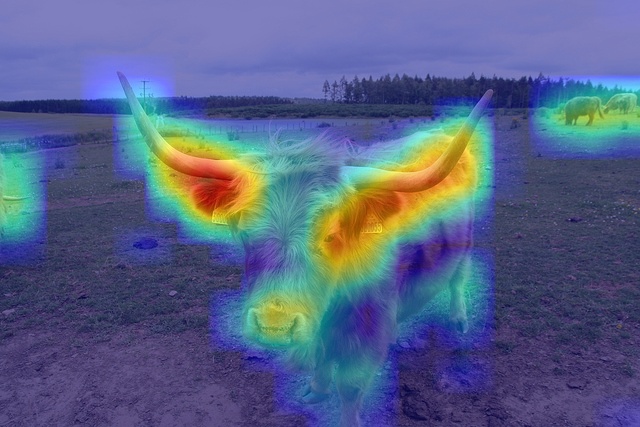}
        \caption{self-attention}\label{fig:detr}
    \end{subfigure}
\end{center}
\caption{
Active regions of CNN-based and DETR-based detectors: (a) Image w/ bounding boxes, (b) ATSS \cite{zhang2020bridging} and (c) Deformable DETR \cite{zhu2020deformable}.}
\label{fig:activate region}
\end{figure}

\paragraph{Issues in feature-level distillation} Most feature-level distillation methods, e.g., \cite{chang2023detrdistill, yang2022focal, guo2021distilling}, focus on the backbone\footnote{In this paper, the term backbone is used loosely to encompass both the traditional backbone network and the neck layers (e.g., FPN).} features that capture only local context from CNNs.
Additionally, previous methods \cite{li2017mimicking, zhang2020improve} often rely on the teacher model in all locations, which can sometimes mislead the student model. Also, these works typically fail to balance the contributions from the foreground and background as well as from objects of different sizes. \citet{ijcai2024p74} focus exclusively on decoder distillation, overlooking the crucial context information embedded in the encoder features. 
%\citet{chang2023detrdistill} utilizes the teacher’s queries to interact with convolution-based FPN features. 
%However, those approaches all fail to leverage the richer global context available in the encoder outputs.
To address these challenges, we propose using location-and-context-aware DETR memory (i.e., transformer encoder output) as a more appropriate alternative for distillation. The encoder processes the sequence of token embeddings, including the positional encoding, through multiple layers of self-attention and feed-forward neural networks. Unlike local CNN features that most previous works use, the memory in DETR captures the transformer-specific global context provided by self-attention, allowing for better reasoning about long-range relationships within the entire image, which is crucial for object detection \cite{hu2018relation}.
Table~\ref{tab:Preliminary} demonstrates the superiority of distilling the memory over distilling backbone features or using a combination of both.
\begin{table}[t]
\centering
    \begin{tabular}{l|ccc}
        \toprule
        Strategy &  AP & $AP_{50}$ & $AP_{75}$ \\
        \midrule
        DINO ResNet-50 (T) & 66.2 & 91.3 & 75.5 \\
        DINO ResNet-18 (S) & 56.1 & 85.1 & 62.9 \\
        Backbone Features  & 57.4 & 85.2   & 63.8 \\
        DETR Memory & \textbf{59.9} & \textbf{87.7} & \textbf{67.5}  \\
        Backbone + DETR Memory & 58.1 & 85.5 & 65.9 \\
        \bottomrule
    \end{tabular}
    \caption{Performance comparison of DINO distillation using different feature representations on the KITTI dataset. ``DETR Memory" refers to the encoder output. Distilling memory outperforms distilling backbone (FPN) features or using a combination of both. T: Teacher, S: Student.}
    \label{tab:Preliminary}
\end{table}
Also, while the DETR self-attention mechanism effectively captures complex dependencies between tokens, it does not inherently alter the sequence order.
As illustrated in the attention maps of Figure \ref{fig:activate region}, both convolutional features and those derived from self-attention maintain a strong indication of spatial locations, with active regions consistently focusing on the foreground.
Motivated by this observation, we propose transforming the ground truth location information to emphasize relevant, contextually-enriched features in the flattened DETR memory during distillation.
%and mitigates inheriting their errors. 
%Notably, we are the first to utilize location-and-context-aware memory for distillation.

\paragraph{Issues in distilling DETR logits}
DETR logit-level distillation faces stability issues due to mismatched distillation points between the teacher and student models. Unlike convolution-based detectors with a fixed spatial arrangement of anchors, DETR's unordered box predictions lack a natural one-to-one correspondence, complicating the alignment between the teacher and student outputs.
\citet{chang2023detrdistill} and \citet{ijcai2024p74} rely on Hungarian matching for teacher-student logit alignment, but this approach results in many meaningless/background predictions being matched unstably.
While \citet{wang2024knowledge} propose consistent distillation points to guide the process, these points are generated randomly or solely based on the fallible teacher model, ignoring the precise location information readily available from the ground truth during distillation. To overcome these challenges, we propose leveraging both category and location information from the ground truth to generate consistent target-aware queries. These queries are designed to be unlearnable and are fed into both the student and teacher models, ensuring that the distillation process precisely and consistently identifies the most informative (and global-context-aware) feature areas for effective knowledge transfer.

In summary, to address the unsolved challenges in DETR distillation at both the feature and logit levels, we propose Consistent Location-and-Context-aware Knowledge Distillation (CLoCKDistill). Our method takes advantage of transformer-specific global context and incorporates ground truth information into both the encoder memory and decoder query for consistent and location-informed distillation. Our contributions are threefold:

\begin{itemize}
    \item We propose global-context-aware feature distillation by utilizing the contextually rich representations in the DETR memory (encoder output). Unlike CNN-generated backbone features, DETR memory is more refined and captures long-range dependencies through self-attention.
    \item Furthermore, we take advantage of the implicit token ordering in DETR memory and utilize ground truth location data to direct feature distillation attention toward more relevant features.
    \item Finally, for logit distillation, we design target-aware decoder queries. These ground-truth-informed queries provide consistent and precise spatial guidance to both the teacher and student models on where to focus in memory when generating logits for distillation.
\end{itemize}

\section{Related Work}
\subsection{Visual Object Detection}
Visual object detection has greatly improved with CNNs, leading to two main types of CNN detectors: two-stage \cite{ren2015faster, cai2018cascade, chu2020detection} and one-stage \cite{redmon2018yolov3, duan2019centernet, tian2019fcos, kong2020foveabox, li2020generalized}.
%Two-stage detectors \cite{cai2018cascade, he2017mask, ren2015faster} generate region proposals before refining and categorizing them, offering higher accuracy at the cost of longer inference times. In contrast, one-stage detectors \cite{lin2017focal, liu2016ssd, redmon2018yolov3, duan2019centernet, tian2019fcos} make predictions directly on feature maps using pre-defined anchor boxes, trading some manual design effort for greater efficiency. 
CNN detectors often use post-processing steps like Non-Maximum Suppression (NMS) to refine final bounding box predictions.
Unlike CNN object detectors, DETR approaches object detection as a set prediction problem using bipartite matching. \citet{carion2020end} introduce the initial end-to-end transformer-based detector without any post-processing.
\citet{dai2021dynamic}, \citet{gao2021fast}, and \citet{sun2021rethinking} attempt to mitigate the slow convergence issue of DETR. 
Deformable DETR \cite{zhu2020deformable} utilizes a deformable attention module that generates a small fixed number of sampling points for each query element, enhancing efficiency by only focusing on relevant feature areas. Additionally, it incorporates prior information (e.g., expected locations of objects) into object queries to further improve detection accuracy. Conditional-DETR \cite{meng2021conditional} separates the content and positional components in object queries, creating positional queries based on the spatial coordinates of objects. DAB-DETR \cite{liu2022dab} further integrates the width and height of bounding boxes into the positional queries. Anchor DETR \cite{wang2022anchor} encodes 2-D anchor points as object queries and designs a row-column decoupled attention to reduce memory cost. DINO \cite{zhang2023dino} presents denoising training methods that utilize noisy ground-truth labels to enhance model training stability and performance. Despite the promising progress, DETRs' complexity still poses challenges for practical use and deployment. Moreover, only a limited number of studies have explored compression techniques on DETRs.

\subsection{Knowledge Distillation for Object Detectors}

The majority of knowledge distillation methods for object detection are designed for CNN-based detectors, utilizing logits \cite{zheng2022localization}, intermediate features \cite{zhang2020improve}, and the relationships between features across different samples \cite{dai2021general}. The first application of KD to object detection by \cite{chen2017learning} involves distilling features from the neck and logits from the classification and regression heads. \citet{li2017mimicking} focus on distilling features from the RPN head. However, not all features are equally useful. 
%\citet{wang2019distilling} introduce masks to concentrate on regions near ground-truth bounding boxes. 
\citet{dai2021general} direct more attention to regions where teacher and student predictions differ most, while \citet{yang2022focal} employ activation-based attention maps to guide feature distillation, incorporating the GcBlock \cite{cao2019gcnet} to capture the relations between pixels.

Despite advancements in knowledge distillation for CNN detectors, most KD techniques are not well-suited for DETRs, with only a few methods tackling the unique challenges they present.
\citet{ijcai2024p74} use Hungarian matching to align queries between the teacher and student models and repurpose the teacher’s queries as an auxiliary group for alignment. However, their approach is limited to decoder-level distillation. \citet{chang2023detrdistill} apply similar matching methods for logit distillation and utilize the teacher’s queries to interact with FPN features, aiming to enhance the distillation of important CNN-based features. Both methods neglect the rich contextual information embedded in the encoder output features.
% \citet{chang2023detrdistill} introduce a distillation method for DETR using Hungarian matching to align queries between the teacher and student models, but it often leads to instability by matching irrelevant queries. 
To tackle the issue of inconsistent distillation points, \citet{wang2024knowledge} employ a predefined set of queries shared between the student and teacher models. However, these queries can sometimes be unreliable, as they are either randomly generated or exclusively derived from the fallible teacher model.
In this paper, we propose using ground truth to design target-aware queries for consistent and more precise logit distillation. Moreover, for more effective feature distillation, instead of distilling backbone features as previous works do, we distill the transformer encoder output (i.e., memory) that captures the global context for each pixel, and employ ground truth location information to emphasize key memory features.
The ability to capture global context is a key advantage of DETRs over CNNs, and we are among the first to explicitly leverage and refine this important knowledge and transfer it to the student model during distillation. 

\section{Methodology}
\begin{figure*}[h]
\begin{center}
\includegraphics[width=0.99\linewidth]{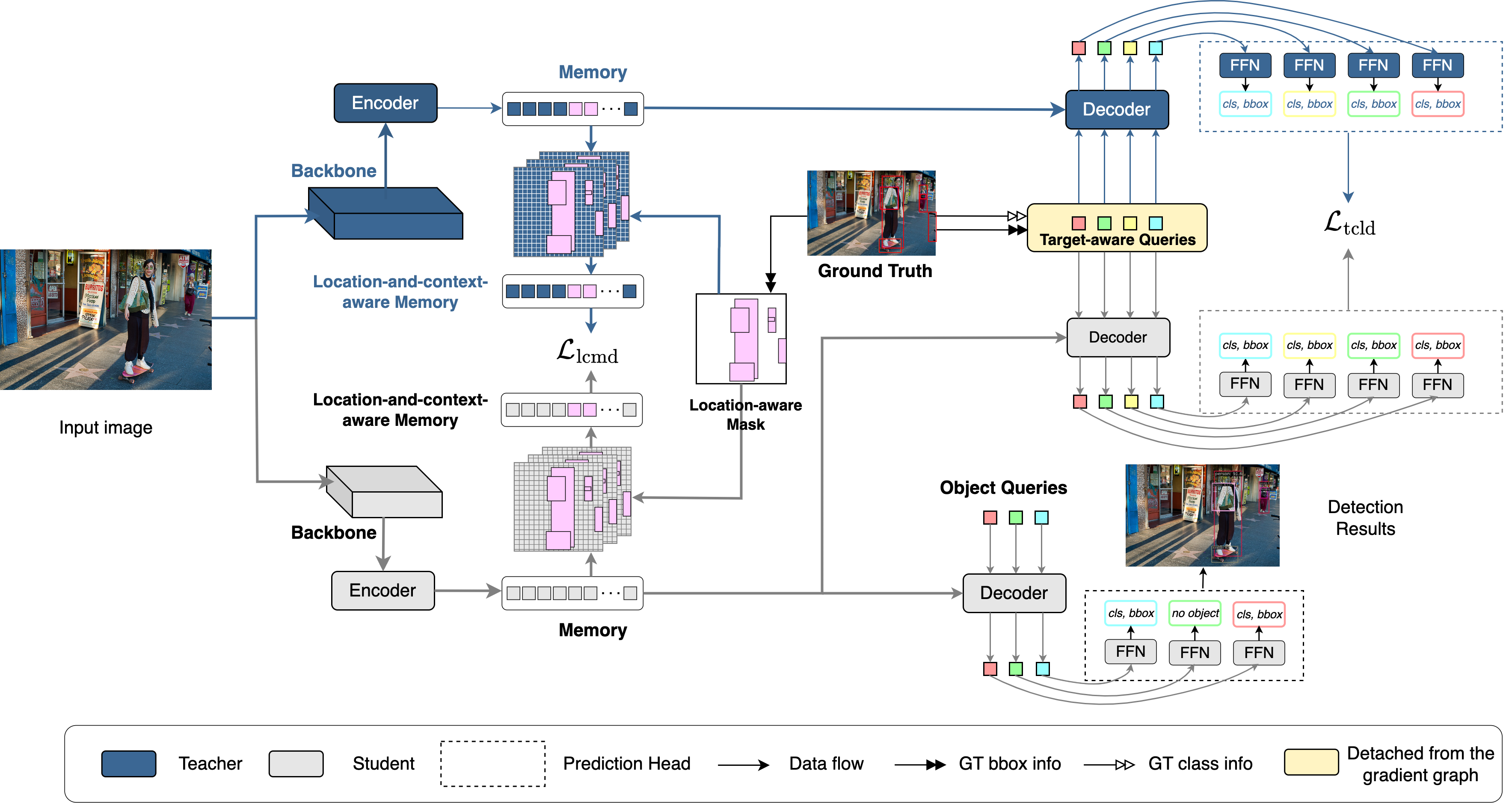}
\end{center}
\caption{Overview of the proposed Consistent Location-and-Context-aware Knowledge Distillation for DETRs.}
\label{fig:overview}
\end{figure*}
DETR has three components: a backbone, an encoder-decoder transformer, and a set of learnable object queries. The backbone generates CNN features $F \in \mathbb{R}^{H\times W\times C}$, where $H$, $W$, and $C$ correspond to the height, width, and number of channels, respectively. These CNN features then pass through the transformer encoder. The output is referred to as DETR memory, denoted as $A \in \mathbb{R}^{HW\times D}$, where $D$ is the embedding size of the transformer. Most existing knowledge distillation works focus on the CNN features from the backbone but overlook the memory. However, we argue that it is the self-attention in the encoder layers that provides each pixel with global context and long-range dependencies, in addition to local feature refinement. Such information helps the model understand the context of an object within the entire image. Therefore, it should be included in the distillation process.
In this paper, we propose to perform distillation on this memory rather than on the backbone features, thereby leveraging the enriched transformer-specific contextual information. Furthermore, we exploit ground truth to emphasize object-related areas in memory, enabling the student detector to focus on the most relevant features without worrying about losing the global context.
On the decoder side, $N$ object queries $Q \in \mathbb{R}^{N\times D}$ first undergo self-attention to interact with each other before proceeding to cross-attention with the encoder's output memory. Instead of utilizing all learnable queries, we introduce target-aware distillation queries that use fixed ground truth information, remaining consistent between the student and teacher models, to ensure precise identification of informative areas during distillation. Each query is subsequently decoded into a bounding box prediction (coordinates $\hat{b}$ and class scores $\hat{c}$) by a feed-forward network (FFN). 
%. Here, $\hat{b}_i = (\hat{b}_{x_i}, \hat{b}_{y_i}, \hat{b}_{w_i}, \hat{b}_{h_i})$, where $i$ refers to the $i$-th query prediction out of $N$, and $(x, y, w, h)$ correspond to the bounding box's center coordinates, width, and height. 
Figure \ref{fig:overview} offers an overview of our CLoCKDistill framework and we will delve into the details in the upcoming subsections.

\subsection{Location-and-context-aware Memory Distillation}
\label{sec:location-and-context-aware Memory Distillation}

Feature-based distillation methods have shown promising performance for both CNN and transformer-based detectors \cite{wang2019distilling, yang2022focal, chang2023detrdistill}. They distill features from convolution layers, such as those in the Feature Pyramid Network (FPN). The objective function for CNN-feature distillation is formulated as:
\begin{equation}
\mathcal{L}_{\text{fea}} = \frac{1}{CHW} \sum_{k=1}^{C} \sum_{i=1}^{H} \sum_{j=1}^{W} \left( F^T_{k,i,j} - f(F^S_{k,i,j}) \right)^2 \,,
\end{equation}
\noindent where $F^T$ and $F^S$ denote the features from the teacher and student models, respectively, and $f$ is an adaptation layer that reshapes $F^S$ to match the dimensions of $F^T$. However, these methods do not distinguish between features of different parts of the image, such as the foreground and the background. More importantly, they fail to capture the global and long-range relationships among different pixels. To address those problems in existing knowledge distillation methods, we propose our Location-and-context-aware memory distillation for DETR models.

We use encoder memory for distillation because it can well capture the global context and long-range dependencies for each pixel through the encoder self-attention mechanism. As demonstrated in Table~\ref{tab:Preliminary}, distilling the memory yields better performance compared to distilling the backbone features. Also, although the encoder memory is flattened, it still retains the spatial order. The $p$-th pixel/point in the encoder memory can be converted to 2-D spatial coordinates through the following conversion:
\begin{equation}
    O(p) = (\left\lfloor \frac{p}{W} \right\rfloor, p\bmod W) \,,
\end{equation}
\noindent where $W$ is the width of the backbone feature map. %$\lfloor \rfloor$ and $mod$ stand for the floor and the modulo operations, respectively.
To direct the student model's attention to more relevant memory areas, we draw upon ground truth location information to differentiate between the foreground and the background. To this end, we design a location-aware mask $M$:
\begin{equation}
M_{p}= \left\{
\begin{array}{cl}
1,\quad & \text{if} \quad O(p)\in \mathcal{G}\\
0,\quad & \text{otherwise}
\end{array} \right. \,,
\end{equation}
\noindent which takes a value of 1 when the converted 2D position of the $p$-th memory point falls within a ground truth bounding box $\mathcal{G}$, and 0 otherwise.

Moreover, we need to balance the contributions from objects of different sizes and consider the varying foreground/background ratios in different images. Larger-scale objects contribute more to the loss because of their greater pixel coverage, which can adversely affect the distillation attention given to smaller objects. 
Also, when the vast majority of pixels in an image are background, the background's influence overwhelms the attention given to foreground objects.
To address these issues, we design a scale mask $S$ as:
\begin{equation}
S_{p} = \left\{
\begin{array}{cl}
\frac{1}{H_k W_k}, & \text{if} \quad O(p) \in BBox_k \\
\frac{1}{N_{bg}}, & \text{otherwise}
\end{array} \right. \,,
\end{equation}
\noindent where $H_k$ and $W_k$ represent the height and width of the $k$-th ground-truth bounding box $BBox_k$, and $N_{bg} = \sum_{i=1}^{H} \sum_{j=1}^{W} (1 - M_{i,j})$. If a pixel belongs to multiple objects, we choose the smallest one to calculate $S$.

We apply the $M$ and $S$ masks to both the teacher and student models' encoder memory, aiming to assist the student in identifying and learning the most relevant and context-aware knowledge from the teacher. Our location-and-context-aware memory distillation loss is defined as:
\begin{equation}\label{eq:lcmd}
\begin{aligned}
    \mathcal{L}_{\text{lcmd}} & = 
      \alpha \sum_{p=1}^{P} \sum_{d=1}^{D} M_{p} S_{p} ( A^T_{p,d} - A^S_{p,d})^2\\
     & + \beta \sum_{p=1}^{P} \sum_{d=1}^{D} (1- M_{p}) S_{p} ( A^T_{p,d} - A^S_{p,d})^2 \,.
\end{aligned}
\end{equation}
\noindent where $A^T$ and $A^S$ denote the memories of the teacher and student detectors, respectively. $P$ is the number of memory points across all locations and scales, and $D$ represents the embedding size of the transformer. $\alpha$ and $\beta$ are balancing hyperparameters. It is worth mentioning that when the backbone CNN features span multiple scales, as in Deformable DETR \cite{zhu2020deformable} and DINO \cite{zhang2023dino}, $P = \sum_{l=1}^L H_l W_l$. Correspondingly, we project $M$ and $S$ to a specific scale level $l$, yielding $M'_l \in \mathbb{R}^{P_l\times 1}$ and $S'_l \in \mathbb{R}^{P_l\times 1}$. These projections are then applied to the memory at that scale $A_l \in \mathbb{R}^{P_l\times D}$ to obtain location-and-context-aware memory for that specific scale. Finally, we concatenate the location-and-context-aware memories from all scales and distill them together.

Our location-and-context-aware memory distillation enhances student learning by leveraging location information from ground truth and rich contextual information available for each pixel in the memory. Both contribute to more effective feature-level knowledge transfer. 

\subsection{Target-aware Consistent Logit Distillation}

Apart from feature distillation, logit distillation has also been shown beneficial for detection tasks \cite{zheng2022localization, wang2024knowledge}. However, in DETR detectors, queries are disordered, resulting in inconsistent prediction matching between the teacher and student models, which hinders effective logit distillation. To address this issue, we propose target-aware distillation queries that incorporate ground truth information to precisely and consistently pinpoint the most informative memory areas for the student to learn from the teacher model. As shown in Figure~\ref{fig:overview}, both the student and teacher models employ the same set of target-aware queries, which remain unlearnable throughout the distillation process.

We utilize both the class label and the position parameters of a ground truth bounding box to generate its corresponding target-aware query for distillation. 
We first employ an embedding layer $\text{Embed}_c$ to convert a category label into a $D$-dimensional vector representation, i.e., content query:
\begin{equation}
    q_{\text{cont}} = \text{Embed}_c(c) \,.
\end{equation}
\noindent Then, we use a multilayer perceptron ($\text{MLP}$) to project the ground truth bounding box parameters (i.e., center coordinates, width, and height) to what we call positional query:
\begin{equation}
    q_{\text{pos}} = \text{MLP}\left(b_{x}, b_{y}, b_{w}, b_{h}\right) \,.
\end{equation}
\noindent Finally, we sum the two $D$-dimensional vectors $q_{cont}$ and $q_{pos}$ to produce our target-aware query $q_{\text{target}}$:
\begin{equation}
q_{\text{target}} = q_{\text{cont}} + q_{\text{pos}} \,.
\end{equation}
It is worth noting that the proposed target-aware queries can be seamlessly combined with other successful queries for enhanced performance. In our experiments, we also include the queries identified as effective in \cite{wang2024knowledge}.
Our distillation queries pass through several stages of the decoder, each involving self-attention and cross-attention mechanisms. Cross-attention enables the queries to focus on relevant parts of our location-and-context-aware memory, before making predictions $\{\hat{\mathbf{c}}, \hat{\mathbf{b}}\}$, where $\hat{\mathbf{c}}$ indicates (unnormalized) class scores or logits, and $\hat{\mathbf{b}}$ stands for corresponding bounding boxes.
Considering all $E$ decoder stages and $G$ distillation points/queries, we define our target-aware consistent logit distillation loss as follows:
\begin{equation}\label{eq:tcld}
\begin{aligned}
\mathcal{L}_{\text{tcld}} & = \sum_{e=1}^E\sum_{g=1}^{G} w_{e,g} [ \lambda_{\text{cls}} \mathcal{L}_{\text{KL}} (\sigma(\frac{\hat{\mathbf{c}}_{e,g}^t}{T}) \parallel \sigma(\frac{\hat{\mathbf{c}}_{e,g}^s}{T})) + \\
&\lambda_{\text{L1}} \mathcal{L}_{\text{L1}} (\hat{\mathbf{b}}_{e,g}^s, \hat{\mathbf{b}}_{e,g}^t) 
+ \lambda_{\text{GIoU}} \mathcal{L}_{\text{GIoU}} (\hat{\mathbf{b}}_{e,g}^s, \hat{\mathbf{b}}_{e,g}^t)],
\end{aligned}
\end{equation}
\noindent where superscripts $t$ and $s$ indicate the teacher and student models, respectively. $\sigma$ stands for the softmax function. Temperature $T$ controls the smoothness of the output distribution. $\mathcal{L}_{\text{KL}}, \mathcal{L}_{\text{L1}}$, and $\mathcal{L}_{\text{GIoU}}$ represent KL-divergence, L1, and GIoU losses, respectively, with $\lambda_{\text{cls}}, \lambda_{\text{L1}}$, and $\lambda_{\text{GIoU}}$ as their corresponding balancing hyperparameters. We define
\begin{equation}
    w_{e,g} = \max_{c \in [0, C]} \sigma (c^t_{e,g})
\end{equation}
\noindent to emphasize distillation queries with higher teacher confidence. 
Adding $\mathcal{L}_{\text{lcmd}}$ and $\mathcal{L}_{\text{tcld}}$ to the original detection loss $\mathcal{L}_{\text{det}}$, we obtain our total training loss for the student model:
\begin{equation}
    \mathcal{L} = \mathcal{L}_{\text{lcmd}} + \mathcal{L}_{\text{tcld}} + \mathcal{L}_{\text{det}} \,.
\end{equation}

\section{Experiments and Results}
\begin{table*}[h]
    \centering
    \addtolength{\tabcolsep}{-0.2pt}
    % \resizebox{\textwidth}{!}{%
    \begin{tabular}{lcccccccccccc}
        \toprule
        Models & Backbone & Epochs & AP & $AP_{50}$ & $AP_{75}$ & $AP_S$ & $AP_M$ & $AP_L$ & GFLOPs & Params \\
        \midrule
        \multirow{4}{*}{DINO} & ResNet-101(T) & 36 & 66.9 & 91.3 & 77.5 & 60.3 & 66.2 & 73.4 & 197 & 66M \\
         & ResNet-50(S) & 12 & 57.7 & 86.4 & 64.7 & 46.1 & 57.6 & 68.0 & 155 & 47M \\
         % & KD-DETR & 12 & 62.1 & 88.7 & 72.3 & 55.2 & 61.8 & 67.9 & 155 & 47M \\
         & Ours & 12 & 63.7 & 89.7 & 72.6 & 57.3 & 63.2 & 69.8 & 155 & 47M \\
         & Gains & - & \textbf{+6.0} & \textbf{+3.3} & \textbf{+7.9} & \textbf{+11.2} & \textbf{+5.6} & \textbf{+1.8} & - & - \\
        \midrule
        \multirow{4}{*}{DINO} & ResNet-50(T) & 36 & 66.2 & 91.3 & 75.5 & 59.5 & 66.0 & 72.6 & 155 & 47M \\
         & ResNet-18(S) & 12 & 56.1 & 85.1 & 62.9 & 45.5 & 55.7 & 65.4 & 128 & 31M\\
         % & KD-DETR & 12 & 60.9 & 88.2 & 68.6 & 53.4 & 60.7 & 67.9 & 128 & 31M \\
         & Ours & 12 & 62.5 & 89.2 & 72.0 & 55.7 & 62.3 & 67.8 & 128 & 31M \\
         & Gains & - & \textbf{+6.4} & \textbf{+4.1} & \textbf{+9.1} & \textbf{+10.2} & \textbf{+6.6} & \textbf{+2.4} & - & - \\
        \midrule
        \multirow{4}{*}{DAB-DETR} & ResNet-101(T) & 50 & 50.9 & 81.6 & 54.1 & 37.7 & 50.8 & 61.9 & 98 & 63M \\
         & ResNet-50(S) & 50 & 47.5 & 80.1 & 48.6 & 30.8 & 47.6 & 62.3 & 56 & 44M \\
         % & KD-DETR & 50 & 48.5 & 81.6 & 50.4 & 34.9 & 48.5 & 60.6 & 56 & 44M \\
         & Ours & 50 & 50.1 & 81.8 & 54.3 & 36.0 & 50.6 & 60.9 & 56 & 44M \\
         & Gains & - & \textbf{+2.6} & \textbf{+1.7} & \textbf{+5.7} & \textbf{+5.2} & \textbf{+3.0} & \textbf{-1.4} & - & - \\
        \midrule
        \multirow{4}{*}{DAB-DETR} & ResNet-50(T) & 50 & 47.5 & 80.1 & 48.6 & 30.8 & 47.6 & 62.3 & 56 & 44M \\
         & ResNet-18(S) & 50 & 39.8 & 74.6 & 38.4 & 21.3 & 39.5 & 57.0 & 31 & 31M\\
         % & KD-DETR & 50 & 44.5 & 77.3 & 45.2 & 29.4 & 43.4 & 60.0 & 31 & 31M \\
         & Ours & 50 & 45.2 & 78.1 & 46.2 & 30.1 & 44.9 & 60.3 & 31 & 31M \\
         & Gains & - & \textbf{+5.4} & \textbf{+3.5} & \textbf{+7.8} & \textbf{+8.8} & \textbf{+5.4} & \textbf{+3.3} & - & - \\
        \midrule
        \multirow{4}{*}{Deformable-DETR} & ResNet-101(T) & 50 & 61.1 & 88.9 & 70.1 & 51.7 & 60.9 & 70.3 & 148 & 59M \\
         & ResNet-50(S) & 50 & 58.7 & 87.6 & 67.3 & 51.3 & 58.3 & 67.7 & 106 & 40M \\
         % & KD-DETR & 50 & 59.9 & 88.1 & 68.7 & 53.3 & 58.8 & 68.1 & 106 & 40M \\
         & Ours & 50 & 60.9 & 88.5 & 68.1 & 53.0 & 59.2 & 68.6 & 106 & 40M \\
         & Gains & - & \textbf{+2.2} & \textbf{+0.9} & \textbf{+0.8} & \textbf{+1.7} & \textbf{+0.9} & \textbf{+0.9} & - & - \\
        \midrule
        \multirow{4}{*}{Deformable-DETR} & ResNet-50(T) & 50 & 58.7 & 89.8 & 66.8 & 51.3 & 58.3 & 69.0 & 106 & 40M \\
         & ResNet-18(S) & 50 & 56.9 & 88.5 & 63.0 & 48.8 & 57.1 & 66.1 & 79 & 24M\\
         % & KD-DETR & 50 & 58.5 & 88.9 & 65.6 & 52.2 & 58.3 & 65.6 & 79 & 24M \\
         & Ours & 50 & 59.8 & 89.1 & 68.8 & 54.8 & 60.9 & 68.2 & 79 & 24M \\
         & Gains & - & \textbf{+2.9} & \textbf{+0.6} & \textbf{+5.8} & \textbf{+6.0} & \textbf{+3.8} & \textbf{+2.1} & - & - \\
        \bottomrule
    \end{tabular}
    \caption{Distillation results of our CLoCKDistill method compared to the baseline across different DETR detectors on the KITTI dataset. T: Teacher, S: Student. Input image dimensions: (402, 1333). All models converged within the indicated epochs.}
    \label{tab:KITTI}
    % }
    \vspace{0.2in}
\end{table*}

\begin{table*}[h]
    \centering
    % \resizebox{\textwidth}{!}{%
    \addtolength{\tabcolsep}{-0.2pt}
    \begin{tabular}{lccccccccccccc}
        \toprule
        Models & Backbone & Epochs & AP & $AP_{50}$ & $AP_{75}$ & $AP_S$ & $AP_M$ & $AP_L$ & GFLOPs & Params \\
        \midrule
        \multirow{4}{*}{DINO} & ResNet-50(T) & 12 & 49.0 & 66.4 & 53.3 & 31.4 & 52.2 & 64.0 & 249 & 47M\\
         & ResNet-18(S) & 12 & 43.5 & 60.7 & 47.1 & 25.3 & 46.1 & 57.6 & 204 & 31M\\
         % & KD-DETR & 12 & 46.1 & 63.1 & 50.1 & 29.2 & 48.7 & 61.1 & 204 & 31M \\
         & Ours & 12 & 46.4 & 63.3 & 50.4 & 28.6 & 48.7 & 61.4 &204 & 31M \\
         & Gains & - & \textbf{+2.9} & \textbf{+2.6} & \textbf{+3.3} & \textbf{+3.3} & \textbf{+2.6} & \textbf{+3.8} & - & -
         \\
        \midrule
        \multirow{4}{*}{DAB-DETR} & ResNet-50(T) & 50 & 42.3 & 63.0 & 45.1 & 21.6 & 46.0 & 61.3 & 92 & 44M\\
         & ResNet-18(S) & 50 & 33.5 & 54.1 & 34.5 & 13.7 & 35.9 & 52.5 & 50 & 31M\\
         % & KD-DETR & 50 & 35.2 & 54.9 & 36.8 & 14.5 & 38.2 & 53.9 &50 & 31M \\
         & Ours & 50 & 36.3 & 56.2 & 37.9 & 15.5 & 39.6 & 55.3 & 50 & 31M \\
         & Gains & - & \textbf{+2.8} & \textbf{+2.1} & \textbf{+3.4} & \textbf{+1.8} & \textbf{+3.7} & \textbf{+2.8} & - & -
         \\
        \midrule
        \multirow{4}{*}{Deformable-DETR} & ResNet-50(T) & 50 & 44.3 & 63.2 & 48.6 & 26.8 & 47.7 & 58.8 & 174 & 40M\\
         & ResNet-18(S) & 50 & 37.8 & 56.7 & 40.8 & 19.4 & 40.5 & 51.7 & 129 & 24M \\
         % & KD-DETR & 50 & 39.0 & 56.4 & 41.8 & 20.9 & 41.6 & 54.0 & 129 & 24M \\
         & Ours & 50 & 40.0 & 57.8 & 43.4 & 21.9 & 42.6 & 54.1 & 129 & 24M \\
         & Gains & - & \textbf{+2.2} & \textbf{+1.1} & \textbf{+2.6} & \textbf{+2.5} & \textbf{+2.1} & \textbf{+2.4} & - & -
         \\
        \bottomrule
    \end{tabular}
    \caption{Distillation results of our CLoCKDistill method compared to the baseline across different DETR detectors on the COCO dataset. T: Teacher, S: Student. Input image dimensions: (1064, 800). All models converged within the indicated epochs.}
    \label{tab:COCO}
    % }
\end{table*}

% \begin{figure*}[h]
% \begin{center}
%     \begin{subfigure}[b]{0.19\textwidth}
%     \includegraphics[]{AnonymousSubmission/Images/000000000154_gt.jpg}
%     \caption{(a)}
%     \end{subfigure}
%     \begin{subfigure}[b]{0.19\textwidth}
%     \includegraphics[]{AnonymousSubmission/Images/teacher_000000000154.jpg}
%     \caption{(b)}
%     \end{subfigure}
% \includegraphics[width=0.19\linewidth]{AnonymousSubmission/Images/student_000000000154.jpg}
% \includegraphics[width=0.19\linewidth]{AnonymousSubmission/Images/kddetr_000000000154.jpg}
% \includegraphics[width=0.19\linewidth]{AnonymousSubmission/Images/ours_000000000154.jpg}
% \end{center}
% \caption{
% Attention maps from (a) Image w/ bounding boxes, (b) the teacher model, (c) the student model (d) KD-DETR distilled \cite{wang2022distilling} (e) our distilled model. Our distilled model allocates more focus to relevant objects, where red denotes the highest attention and blue the lowest.}
% \label{fig:activate vis}
% \end{figure*}
\begin{figure*}
\begin{center}
    \begin{subfigure}[b]{0.19\textwidth}
        \includegraphics[width=\textwidth]{sec/Images/000000000154.jpg}
        \caption{Original Image}\label{fig:ori_zebra}
    \end{subfigure}
    \begin{subfigure}[b]{0.19\textwidth}
        \includegraphics[width=\textwidth]{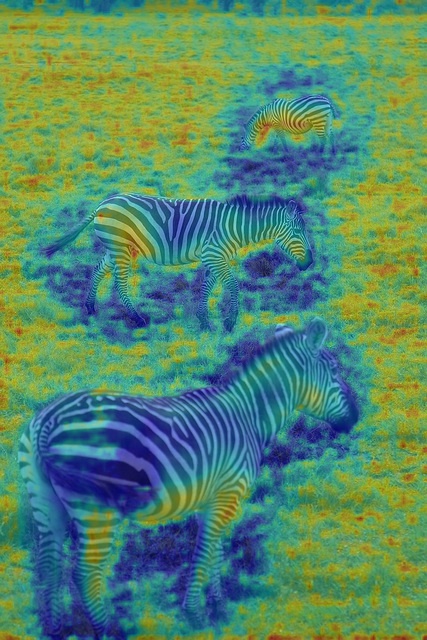}
        \caption{Teacher}\label{fig:teacherattn_zebra}
    \end{subfigure}
    \begin{subfigure}[b]{0.19\textwidth}
        \includegraphics[width=\textwidth]{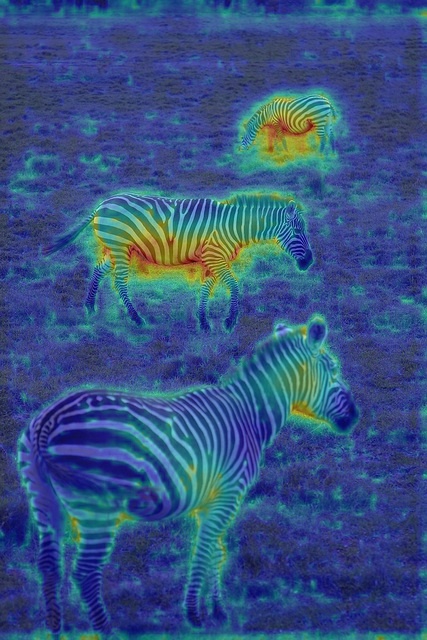}
        \caption{Student baseline}\label{fig:studentattn_zebra}
    \end{subfigure}
    \begin{subfigure}[b]{0.19\textwidth}
        \includegraphics[width=\textwidth]{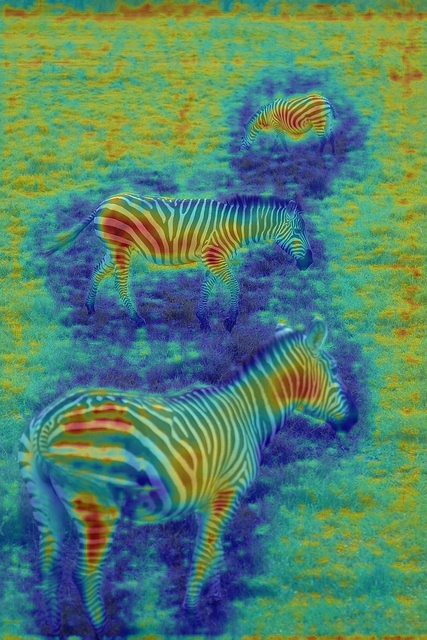}
        \caption{KD-DETR}\label{fig:kddetrattn_zebra}
    \end{subfigure}
    \begin{subfigure}[b]{0.19\textwidth}
        \includegraphics[width=\textwidth]{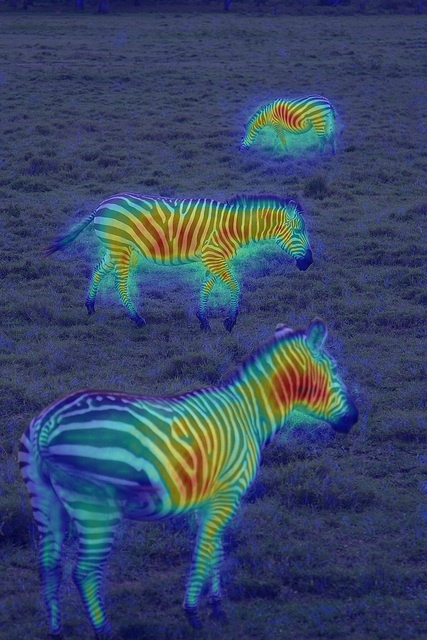}
        \caption{Ours}\label{fig:ourattn_zebra}
    \end{subfigure}

    % \begin{subfigure}[b]{0.19\textwidth}
    %     \includegraphics[width=\textwidth]{sec/Images/000000000143.jpg}
    %     \caption{Original Image}\label{fig:ori_birds}
    % \end{subfigure}
    % \begin{subfigure}[b]{0.19\textwidth}
    %     \includegraphics[width=\textwidth]{sec/Images/teacher_000000000143.jpg}
    %     \caption{Teacher}\label{fig:teacherattn_birds}
    % \end{subfigure}
    % \begin{subfigure}[b]{0.19\textwidth}
    %     \includegraphics[width=\textwidth]{sec/Images/student_000000000143.jpg}
    %     \caption{Student baseline}\label{fig:studentattn_birds}
    % \end{subfigure}
    % \begin{subfigure}[b]{0.19\textwidth}
    %     \includegraphics[width=\textwidth]{sec/Images/kddetr_000000000143.jpg}
    %     \caption{KD-DETR}\label{fig:kddetrattn_birds}
    % \end{subfigure}
    % \begin{subfigure}[b]{0.19\textwidth}
    %     \includegraphics[width=\textwidth]{sec/Images/ours_000000000143.jpg}
    %     \caption{Ours}\label{fig:ourattn_birds}
    % \end{subfigure}
    
\end{center}
\caption{Attention maps of (a) an example COCO image from the (b) teacher, (c) student baseline, (d) KD-DETR distilled \cite{wang2024knowledge}, and (e) our distilled models. Our distilled model allocates more focus to relevant objects and exhibits more clearly defined boundaries. Red indicates the highest attention and blue the lowest.}
\label{fig:activate vis}
\end{figure*}

\subsection{Datasets}
We demonstrate our method's efficacy using two datasets:

\noindent\textbf{KITTI} \cite{Geiger2012CVPR} offers a widely used 2D object detection benchmark for autonomous driving and computer vision research. It specifically represents scenarios that demand low latency and constrained computational resources. The dataset includes seven types of road objects and contains 7,481 annotated images, which are split into a training set and a validation set using an 8:2 ratio. We follow the convention in \cite{lan2024gradient} for pre-processing.
%Specifically, \textit{car, van, truck, tram} are regrouped as Car, \textit{pedestrian, person} as Pedestrian, and \textit{cyclist} as Cyclist.

\noindent\textbf{MS COCO} is a dataset that includes 80 categories spanning a wide range of objects, with 117K training images and 5K validation images.
We adopt COCO-style evaluation metrics, e.g., mean Average Precision (mAP), for both datasets. 
% delete when needed
The validation process for each dataset closely follows these standard metrics to ensure consistency and comparability.

\subsection{Implementation Details}
We evaluate our approach on three state-of-the-art DETR detectors: DAB-DETR \cite{liu2022dab}, Deformable-DETR \cite{zhu2020deformable}, and DINO \cite{zhang2023dino}. All experiments are conducted using the MMDetection framework \cite{mmdetection} on Nvidia A100 GPUs, adhering to the original hyperparameter settings and optimizer configurations. ResNets \cite{he2016deep} are used as backbones for both the teacher and student models. The distillation loss coefficients are set as follows: $\alpha = 5\times 10^{-5}, \beta = 1\times 10^{-7}, \lambda_{kl} = 1$, $\lambda_{L1} = 5$, and $\lambda_{GIoU} = 2$. The number of distillation points is set to 300 for DAB-DETR and Deformable DETR, and 900 for DINO. The number of copies for target-aware queries is set to 3 for all experiments. 

\subsection{Quantitative Results}
Table \ref{tab:KITTI} presents the results on KITTI, highlighting the effectiveness of our distillation method across various DETR models and backbones. Notably, our distillation approach boosts the performance of DINO with ResNet-18 and ResNet-50 backbones by 6.4\% and 6.0\% mAP, respectively. %The student distilled by our method surpasses the state-of-the-art KD-DETR \cite{wang2024knowledge} by 1.6\% for both ResNet-50 and ResNet-18 backbones. 
For DAB-DETR, our method improves the performance of the detectors using ResNet-18 and ResNet-50 backbones by 5.4\% and 2.6\% mAP, respectively. Furthermore, in the case of Deformable DETR, our approach achieves mAP improvements of 2.9\% and 2.2\% for the ResNet-18 and ResNet-50 backbones, respectively.

%We also beat the baseline and competing methods by clear margins on COCO (Tab. \ref{tab:COCO}). For instance, our method boosts DAB-DETR performance by 2.8\% mAP when using the ResNet-18 backbone. This outperforms the cutting-edge distillation method \cite{wang2024knowledge} by 1.1\% mAP.

We also beat the baselines by clear margins on the COCO dataset (Table \ref{tab:COCO}). For instance, our method boosts DINO performance by 2.9\% mAP when using the ResNet-18 backbone. %This outperforms the cutting-edge distillation method \cite{wang2024knowledge} by 1.1\% mAP.

In Table~\ref{tab:kd methods}, we present a comparison of our method with state-of-the-art knowledge distillation methods for object detectors, including FitNet \cite{romero2014fitnets}, FGD \cite{yang2022focal}, MGD \cite{yang2022masked}, and KD-DETR \cite{wang2024knowledge}. Notably, KD-DETR \cite{wang2024knowledge} (CVPR 2024) represents the latest advancement specifically tailored for DETR distillation. According to the results, while KD-DETR demonstrates superior performance over conventional KD methods, the margin over MGD is not significant. Our method enlarges the performance gap between DETR-oriented and conventional KD methods by taking advantage of transformer-specific global context and strategically incorporating ground truth information into both feature and logit distillation. 
% \begin{table*}[h]
% \centering
% \begin{tabular}{|lccccccc|}
% \toprule
% Method & Epoch & AP & $AP_{50}$ & $AP_{75}$ & $AP_s$ & $AP_m$ & $AP_l$ \\ \midrule
% DINO-R50 & 36 & 66.2 & 91.3 & 75.5 & 59.5 & 66.0 & 72.6 \\ 
% DINO-R18 & 12 & 56.1 & 85.1 & 62.9 & 45.5 & 55.7 & 65.4 \\ 
% FGD & 12 & 57.1 & 85.3 & 64.5 & 49.1 & 56.9 & 63.1 \\ 
% MGD & 12 & 60.1 & 87.7 & 67.5 & 53.7 & 53.7 & 59.6 \\ 
% FitNet & 12 & 57.4 & 85.2 & 63.8 & 48.5 & 57.2& 64.9 \\
% KD-DETR & 12 & 60.9 & 88.2 & 68.6 & 53.4 & 60.7 & 67.9 \\ 
% Ours & 12 & \textbf{62.5} & 89.2 & 72.0 & 55.7 & 62.3 & 67.8 \\ \bottomrule
% \end{tabular}
% \caption{Comparison with different state-of-the-art KD methods on DINO.}
% \label{tab:kd methods}
% \end{table*}

\begin{table}[h]
\centering
\begin{tabular}{c|cccc}
\toprule
Method & Epoch & AP & $AP_{50}$ & $AP_{75}$ \\ \midrule
DINO ResNet-50(T) & 36 & 66.2 & 91.3 & 75.5  \\ 
DINO ResNet-18(S) & 12 & 56.1 & 85.1 & 62.9  \\ 
FitNet & 12 & 57.4 & 85.2 & 63.8 \\
FGD & 12 & 57.1 & 85.3 & 64.5 \\ 
MGD & 12 & 60.1 & 87.7 & 67.5 \\ 
KD-DETR & 12 & 60.9 & 88.2 & 68.6  \\ \midrule
\textbf{Ours} & 12 & \textbf{62.5} & \textbf{89.2} & \textbf{72.0} \\ \bottomrule
\end{tabular}
\caption{Comparison of our method with state-of-the-art KD methods. T: Teacher, S: Student. Detector: DINO, Dataset: KITTI. All models converged within the indicated epochs.}
\label{tab:kd methods}
\end{table}

\subsection{Qualitative Results}
The attention maps of different approaches in Figure~\ref{fig:activate vis} also demonstrate our CLoCKDistill's effectiveness. In Figure~\ref{fig:teacherattn_zebra}, the teacher model's attention map reveals wide dispersion, with attention spread across both the foreground objects (zebras) and the background. This broad focus reflects the teacher model's strong representational capacity due to its larger size; however, it also leads to substantial attention on irrelevant background areas, which can complicate or interfere with accurate detection.
Unlike the over-parameterized teacher model capturing many unnecessary and irrelevant features, the capacity-limited student baseline learns to focus most of its attention on the foreground objects while trying to achieve a satisfactory detection performance on its own (as shown in Figure~\ref{fig:studentattn_zebra}). However, this focus is confined to specific boundary areas, and the student baseline exhibits variability in attention across zebras of different scales, highlighting the model's sensitivity to scale changes.
When learning from the teacher via KD-DETR (Figure~\ref{fig:kddetrattn_zebra}), the student model inherits much of the teacher’s irrelevant focus on distracting background features, which impedes its ability to concentrate effectively on objects of interest.
In contrast, our CLoCKDistill model (Figure~\ref{fig:ourattn_zebra}) achieves a more targeted focus, prioritizing relevant objects. Additionally, the attention boundaries are more sharply defined, indicating that our distilled model has developed a clearer and more precise understanding of the objects.

\subsection{Ablation Study}
In this section, we conduct ablation studies on the key components of our CLoCKDistill method and the number of transformer encoder-decoder layers.
%The experiments are on the KITTI dataset using DINO, with ResNet-50 as the teacher and ResNet-18 as the student backbone.
\subsubsection{CLoCKDistill components}
Table \ref{tab:Ablation} presents the influence of the main components of our CLoCKDistill, i.e., distilling memory, masking memory with location info, and target-aware queries. As we can see, memory distillation alone boosts the student model's performance by 3.8\% mAP. Further masking memory with location info enhances performance by 5.5\% mAP. By integrating all components, including target-aware queries, we achieve a 6.4\% mAP improvement for the DINO detector with a ResNet-18 backbone.
% \begin{table*}
%     \centering
%     \addtolength{\tabcolsep}{-0.3pt}
%     \begin{tabular}{lccc|ccccccc}
%         \toprule
%         Models & Memory & Location-aware masks & Target-aware query & AP & $AP_{50}$ & $AP_{75}$ & $AP_S$ & $AP_M$ & $AP_L$ \\
%         \midrule
%         DINO-R18    & & & & 56.1 & 85.1 & 62.9 & 45.5 & 55.7 & 65.4\\
%         DINO-R50-R18 & \checkmark & & & 59.9 & 87.7 & 67.5 & 52.0 & 60.1 & 65.9 \\
%         DINO-R50-R18 & \checkmark & \checkmark & & 61.6 & 89.4 & 68.7 & 54.7 & 61.6 & 67.7 \\
%         DINO-R50-R18 & \checkmark & \checkmark & \checkmark & \textbf{62.5} & \textbf{89.2} & \textbf{72.0} & \textbf{55.7} & \textbf{62.3} & \textbf{67.8} \\
%         \bottomrule
%     \end{tabular}
%     \caption{Ablation study of different CLoCKDistill components. Detector: DINO, dataset: KITTI.}
%     \label{tab:Ablation}
% \end{table*}
\begin{table}[h]
    \centering
    %\addtolength{\tabcolsep}{-0.3pt}
    \begin{tabular}{l|ccc}
        \toprule
        Components & AP & $AP_{50}$ & $AP_{75}$ \\
        \midrule
        Baseline       & 56.1 & 85.1 & 62.9 \\
        w/ Mem        & 59.9 \textbf{(\textuparrow 3.8)} & 87.7 & 67.5  \\
        w/ Mem + LM   & 61.6 \textbf{(\textuparrow 5.5)} & 89.4 & 68.7 \\
        w/ Mem + LM + TQ & \textbf{62.5} \textbf{(\textuparrow 6.4)} & \textbf{89.2} & \textbf{72.0} \\
        \bottomrule
    \end{tabular}
    \caption{Ablation study of different CLoCKDistill components. Mem represents Memory distillation, LM indicates Location-aware Mask, and TQ stands for Target-aware Query. The experiments are conducted on the KITTI dataset using DINO detectors, with ResNet-50 as the teacher and ResNet-18 as the student backbone.}
    \label{tab:Ablation}
\end{table}

\subsubsection{Number of transformer layers}
\begin{comment}
\begin{table}[h]
    \centering
    \addtolength{\tabcolsep}{-0.1pt}
    \begin{tabular}{c|ccc|cc}
        \toprule
        Enc/Dec & AP & $AP_{50}$ & $AP_{75}$ & FPS & Params \\
        \midrule
        6/6 & 56.1 & 85.1 & 62.9 & 27.5 & 31M\\
        Ours & 62.5(+6.4) & 89.2 & 72.0 & 27.5  & 31M\\
        3/6 & 56.1 & 84.8 & 63.4 & 34.7 & 27M\\
        Ours & 60.4(+4.3) & 89.1 & 68.2 & 34.7 & 27M\\
        6/3 & 54.7 & 81.7 & 61.4 & 33.6 & 26M\\
        Ours & 61.7(+5.0) & 88.2 & 70.9 & 33.6 & 26M\\
        3/3 & 52.5 & 78.7 & 59.7 & 44.8 & 22M\\
        Ours & 59.5(+7.0) & 87.2 & 67.2 & 44.8 & 22M \\
        \bottomrule
    \end{tabular}
    \caption{Distillation performance with varying numbers of transformer layers in DINO on the KITTI dataset. Teacher backbone: ResNet-50, student backbone: ResNet-18.}
    \label{tab:enc_dec}
\end{table}
\end{comment}
\begin{table}
    \centering
    \addtolength{\tabcolsep}{-0.1pt}
    \begin{tabular}{c|c|ccc|c}
        \toprule
        Model & Enc/Dec & AP & $AP_{50}$ & $AP_{75}$ & FPS \\
        \midrule
        StuBase & 6/6 & 56.1 & 85.1 & 62.9 & \multirow{2}{*}{27.5}\\
        Ours & 6/6 & 62.5 \textbf{(\textuparrow 6.4)} & 89.2 & 72.0 & \\
        \midrule
        StuBase & 3/6 & 56.1 & 84.8 & 63.4 & \multirow{2}{*}{34.7}\\
        Ours & 3/6 & 60.4 \textbf{(\textuparrow 4.3)} & 89.1 & 68.2 & \\
        \midrule
        StuBase & 6/3 & 54.7 & 81.7 & 61.4 & \multirow{2}{*}{33.6}\\
        Ours & 6/3 & 61.7 \textbf{(\textuparrow 5.0)} & 88.2 & 70.9 &\\
        \midrule
        StuBase & 3/3 & 52.5 & 78.7 & 59.7 & \multirow{2}{*}{44.8}\\
        Ours & 3/3 & 59.5 \textbf{(\textuparrow 7.0)} & 87.2 & 67.2 & \\
        \bottomrule
    \end{tabular}
    \caption{Distillation performance with varying numbers of transformer encoder and/or decoder layers in the DINO detector on the KITTI dataset. StuBase indicates the student baseline with a ResNet-18 backbone. The teacher has a ResNet-50 backbone with 6 encoders and 6 decoders, with an FPS of 19.8. FPS stands for frames per second, which is measured on a single Nvidia A100 GPU.}
    \label{tab:enc_dec}
\end{table}
The transformer detection head requires considerable computation, so reducing the number of encoder and decoder layers can improve model efficiency. To explore this, we conducted experiments with different reduced layer configurations. To address the layer mismatch between the student and teacher models, we applied our distillation method only to the final encoder and decoder layers.

Table~\ref{tab:enc_dec} shows the impact of varying the number of transformer encoder and decoder layers. As expected, reducing the number of encoder/decoder layers decreases the model performance. The student baseline is more sensitive to reductions in decoder layers, while our distilled model is more affected by reductions in encoder layers.
Notably, with only half of the encoder and decoder layers, our distilled model still surpasses the full-scale baseline by 3.4\% in mAP and achieves a 1.6x increase in frames per second (FPS). 

\section{Conclusion}

In this paper, we introduce CLoCKDistill, a novel knowledge distillation method designed for compressing DETR detectors. Our approach incorporates both feature distillation and logit distillation. For feature distillation, we effectively transfer the transformer-specific global context from the teacher to the student by distilling the DETR memory, with the guidance from ground truth. For logit distillation, we propose target-aware queries that provide consistent and precise spatial guidance for both the teacher and student models on where to focus in memory during logit generation. Experimental results on the KITTI and COCO datasets show that our CLoCKDistill significantly enhances the performance of various DETR models, outperforming state-of-the-art methods. Additionally, our method clearly improves focus on relevant objects, as demonstrated by the qualitative analysis of attention maps. The clear attention boundaries suggest that our method could be promising for segmentation tasks, a potential avenue for future research.
{
    \small
    \bibliographystyle{ieeenat_fullname}
    \bibliography{main}

\begin{thebibliography}{40}
\providecommand{\natexlab}[1]{#1}
\providecommand{\url}[1]{\texttt{#1}}
\expandafter\ifx\csname urlstyle\endcsname\relax
  \providecommand{\doi}[1]{doi: #1}\else
  \providecommand{\doi}{doi: \begingroup \urlstyle{rm}\Url}\fi

\bibitem[Cai and Vasconcelos(2018)]{cai2018cascade}
Zhaowei Cai and Nuno Vasconcelos.
\newblock Cascade r-cnn: Delving into high quality object detection.
\newblock In \emph{Proceedings of the IEEE/CVF Conference on Computer Vision and Pattern Recognition}, pages 6154--6162, 2018.

\bibitem[Cao et~al.(2019)Cao, Xu, Lin, Wei, and Hu]{cao2019gcnet}
Yue Cao, Jiarui Xu, Stephen Lin, Fangyun Wei, and Han Hu.
\newblock Gcnet: Non-local networks meet squeeze-excitation networks and beyond.
\newblock In \emph{Proceedings of the IEEE/CVF International Conference on Computer Vision Workshops}, pages 0--0, 2019.

\bibitem[Carion et~al.(2020)Carion, Massa, Synnaeve, Usunier, Kirillov, and Zagoruyko]{carion2020end}
Nicolas Carion, Francisco Massa, Gabriel Synnaeve, Nicolas Usunier, Alexander Kirillov, and Sergey Zagoruyko.
\newblock End-to-end object detection with transformers.
\newblock In \emph{Proceedings of the European Conference on Computer Vision}, pages 213--229. Springer, 2020.

\bibitem[Chang et~al.(2023)Chang, Wang, Xu, Chen, Yang, and Zhao]{chang2023detrdistill}
Jiahao Chang, Shuo Wang, Hai-Ming Xu, Zehui Chen, Chenhongyi Yang, and Feng Zhao.
\newblock Detrdistill: A universal knowledge distillation framework for detr-families.
\newblock In \emph{Proceedings of the IEEE/CVF International Conference on Computer Vision}, pages 6898--6908, 2023.

\bibitem[Chectionen et~al.(2019)Chectionen, Wang, Pang, Cao, Xiong, Li, Sun, Feng, Liu, Xu, Zhang, Cheng, Zhu, Cheng, Zhao, Li, Lu, Zhu, Wu, Dai, Wang, Shi, Ouyang, Loy, and Lin]{mmdetection}
Kai Chectionen, Jiaqi Wang, Jiangmiao Pang, Yuhang Cao, Yu Xiong, Xiaoxiao Li, Shuyang Sun, Wansen Feng, Ziwei Liu, Jiarui Xu, Zheng Zhang, Dazhi Cheng, Chenchen Zhu, Tianheng Cheng, Qijie Zhao, Buyu Li, Xin Lu, Rui Zhu, Yue Wu, Jifeng Dai, Jingdong Wang, Jianping Shi, Wanli Ouyang, Chen~Change Loy, and Dahua Lin.
\newblock {MMDetection}: Open mmlab detection toolbox and benchmark.
\newblock \emph{arXiv preprint arXiv:1906.07155}, 2019.

\bibitem[Chen et~al.(2017)Chen, Choi, Yu, Han, and Chandraker]{chen2017learning}
Guobin Chen, Wongun Choi, Xiang Yu, Tony Han, and Manmohan Chandraker.
\newblock Learning efficient object detection models with knowledge distillation.
\newblock \emph{Advances in Neural Information Processing Systems}, 30, 2017.

\bibitem[Chen et~al.(2024)Chen, Chen, Liu, Tang, and Zeng]{ijcai2024p74}
Xiaokang Chen, Jiahui Chen, Yan Liu, Jiaxiang Tang, and Gang Zeng.
\newblock D3etr: Decoder distillation for detection transformer.
\newblock In \emph{Proceedings of the Thirty-Third International Joint Conference on Artificial Intelligence, {IJCAI-24}}, pages 668--676. International Joint Conferences on Artificial Intelligence Organization, 2024.
\newblock Main Track.

\bibitem[Chu et~al.(2020)Chu, Zheng, Zhang, and Sun]{chu2020detection}
Xuangeng Chu, Anlin Zheng, Xiangyu Zhang, and Jian Sun.
\newblock Detection in crowded scenes: One proposal, multiple predictions.
\newblock In \emph{Proceedings of the IEEE/CVF Conference on Computer Vision and Pattern Recognition}, pages 12214--12223, 2020.

\bibitem[Dai et~al.(2021{\natexlab{a}})Dai, Chen, Yang, Zhang, Yuan, and Zhang]{dai2021dynamic}
Xiyang Dai, Yinpeng Chen, Jianwei Yang, Pengchuan Zhang, Lu Yuan, and Lei Zhang.
\newblock Dynamic detr: End-to-end object detection with dynamic attention.
\newblock In \emph{Proceedings of the IEEE/CVF International Conference on Computer Vision}, pages 2988--2997, 2021{\natexlab{a}}.

\bibitem[Dai et~al.(2021{\natexlab{b}})Dai, Jiang, Wu, Bao, Wang, Liu, and Zhou]{dai2021general}
Xing Dai, Zeren Jiang, Zhao Wu, Yiping Bao, Zhicheng Wang, Si Liu, and Erjin Zhou.
\newblock General instance distillation for object detection.
\newblock In \emph{Proceedings of the IEEE/CVF Conference on Computer Vision and Pattern Recognition}, pages 7842--7851, 2021{\natexlab{b}}.

\bibitem[Duan et~al.(2019)Duan, Bai, Xie, Qi, Huang, and Tian]{duan2019centernet}
Kaiwen Duan, Song Bai, Lingxi Xie, Honggang Qi, Qingming Huang, and Qi Tian.
\newblock Centernet: Keypoint triplets for object detection.
\newblock In \emph{Proceedings of the IEEE/CVF International Conference on Computer Vision}, pages 6569--6578, 2019.

\bibitem[Gao et~al.(2021)Gao, Zheng, Wang, Dai, and Li]{gao2021fast}
Peng Gao, Minghang Zheng, Xiaogang Wang, Jifeng Dai, and Hongsheng Li.
\newblock Fast convergence of detr with spatially modulated co-attention.
\newblock In \emph{Proceedings of the IEEE/CVF International Conference on Computer Vision}, pages 3621--3630, 2021.

\bibitem[Geiger et~al.(2012)Geiger, Lenz, and Urtasun]{Geiger2012CVPR}
Andreas Geiger, Philip Lenz, and Raquel Urtasun.
\newblock Are we ready for autonomous driving? the kitti vision benchmark suite.
\newblock In \emph{Proceedings of the IEEE/CVF Conference on Computer Vision and Pattern Recognition}, 2012.

\bibitem[Guo et~al.(2021)Guo, Han, Wang, Wu, Chen, Xu, and Xu]{guo2021distilling}
Jianyuan Guo, Kai Han, Yunhe Wang, Han Wu, Xinghao Chen, Chunjing Xu, and Chang Xu.
\newblock Distilling object detectors via decoupled features.
\newblock In \emph{Proceedings of the IEEE/CVF Conference on Computer Vision and Pattern Recognition}, pages 2154--2164, 2021.

\bibitem[He et~al.(2016)He, Zhang, Ren, and Sun]{he2016deep}
Kaiming He, Xiangyu Zhang, Shaoqing Ren, and Jian Sun.
\newblock Deep residual learning for image recognition.
\newblock In \emph{Proceedings of the IEEE/CVF Conference on Computer Vision and Pattern Recognition}, pages 770--778, 2016.

\bibitem[Hinton et~al.(2015)Hinton, Vinyals, and Dean]{hinton2015distilling}
Geoffrey Hinton, Oriol Vinyals, and Jeff Dean.
\newblock Distilling the knowledge in a neural network.
\newblock \emph{arXiv preprint arXiv:1503.02531}, 2015.

\bibitem[Hosang et~al.(2017)Hosang, Benenson, and Schiele]{hosang2017learning}
Jan Hosang, Rodrigo Benenson, and Bernt Schiele.
\newblock Learning non-maximum suppression.
\newblock In \emph{Proceedings of the IEEE/CVF Conference on Computer Vision and Pattern Recognition}, pages 4507--4515, 2017.

\bibitem[Hu et~al.(2018)Hu, Gu, Zhang, Dai, and Wei]{hu2018relation}
Han Hu, Jiayuan Gu, Zheng Zhang, Jifeng Dai, and Yichen Wei.
\newblock Relation networks for object detection.
\newblock In \emph{Proceedings of the IEEE/CVF Conference on Computer Vision and Pattern Recognition}, pages 3588--3597, 2018.

\bibitem[Kong et~al.(2020)Kong, Sun, Liu, Jiang, Li, and Shi]{kong2020foveabox}
Tao Kong, Fuchun Sun, Huaping Liu, Yuning Jiang, Lei Li, and Jianbo Shi.
\newblock Foveabox: Beyound anchor-based object detection.
\newblock \emph{IEEE Transactions on Image Processing}, 29:\penalty0 7389--7398, 2020.

\bibitem[Lan and Tian(2024)]{lan2024gradient}
Qizhen Lan and Qing Tian.
\newblock Gradient-guided knowledge distillation for object detectors.
\newblock In \emph{Proceedings of the IEEE/CVF Winter Conference on Applications of Computer Vision}, pages 424--433, 2024.

\bibitem[Li et~al.(2017)Li, Jin, and Yan]{li2017mimicking}
Quanquan Li, Shengying Jin, and Junjie Yan.
\newblock Mimicking very efficient network for object detection.
\newblock In \emph{Proceedings of the IEEE/CVF Conference on Computer Vision and Pattern Recognition}, pages 6356--6364, 2017.

\bibitem[Li et~al.(2020)Li, Wang, Wu, Chen, Hu, Li, Tang, and Yang]{li2020generalized}
Xiang Li, Wenhai Wang, Lijun Wu, Shuo Chen, Xiaolin Hu, Jun Li, Jinhui Tang, and Jian Yang.
\newblock Generalized focal loss: Learning qualified and distributed bounding boxes for dense object detection.
\newblock \emph{Advances in Neural Information Processing Systems}, 33:\penalty0 21002--21012, 2020.

\bibitem[Lin et~al.(2017)Lin, Goyal, Girshick, He, and Doll{\'a}r]{lin2017focal}
Tsung-Yi Lin, Priya Goyal, Ross Girshick, Kaiming He, and Piotr Doll{\'a}r.
\newblock Focal loss for dense object detection.
\newblock In \emph{Proceedings of the IEEE/CVF International Conference on Computer Vision}, pages 2980--2988, 2017.

\bibitem[Liu et~al.(2022)Liu, Li, Zhang, Yang, Qi, Su, Zhu, and Zhang]{liu2022dab}
Shilong Liu, Feng Li, Hao Zhang, Xiao Yang, Xianbiao Qi, Hang Su, Jun Zhu, and Lei Zhang.
\newblock Dab-detr: Dynamic anchor boxes are better queries for detr.
\newblock In \emph{Proceedings of the International Conference on Learning Representations}, 2022.

\bibitem[Meng et~al.(2021)Meng, Chen, Fan, Zeng, Li, Yuan, Sun, and Wang]{meng2021conditional}
Depu Meng, Xiaokang Chen, Zejia Fan, Gang Zeng, Houqiang Li, Yuhui Yuan, Lei Sun, and Jingdong Wang.
\newblock Conditional detr for fast training convergence.
\newblock In \emph{Proceedings of the IEEE/CVF International Conference on Computer Vision}, pages 3651--3660, 2021.

\bibitem[Redmon and Farhadi(2018)]{redmon2018yolov3}
Joseph Redmon and Ali Farhadi.
\newblock Yolov3: An incremental improvement.
\newblock \emph{arXiv preprint arXiv:1804.02767}, 2018.

\bibitem[Ren et~al.(2015)Ren, He, Girshick, and Sun]{ren2015faster}
Shaoqing Ren, Kaiming He, Ross Girshick, and Jian Sun.
\newblock Faster r-cnn: Towards real-time object detection with region proposal networks.
\newblock \emph{Advances in Neural Information Processing Systems}, 28, 2015.

\bibitem[Romero et~al.(2014)Romero, Ballas, Kahou, Chassang, Gatta, and Bengio]{romero2014fitnets}
Adriana Romero, Nicolas Ballas, Samira~Ebrahimi Kahou, Antoine Chassang, Carlo Gatta, and Yoshua Bengio.
\newblock Fitnets: Hints for thin deep nets.
\newblock \emph{arXiv preprint arXiv:1412.6550}, 2014.

\bibitem[Sun et~al.(2021)Sun, Cao, Yang, and Kitani]{sun2021rethinking}
Zhiqing Sun, Shengcao Cao, Yiming Yang, and Kris~M Kitani.
\newblock Rethinking transformer-based set prediction for object detection.
\newblock In \emph{Proceedings of the IEEE/CVF International Conference on Computer Vision}, pages 3611--3620, 2021.

\bibitem[Tian et~al.(2019)Tian, Shen, Chen, and He]{tian2019fcos}
Zhi Tian, Chunhua Shen, Hao Chen, and Tong He.
\newblock Fcos: Fully convolutional one-stage object detection.
\newblock In \emph{Proceedings of the IEEE/CVF International Conference on Computer Vision}, pages 9627--9636, 2019.

\bibitem[Wang et~al.(2019)Wang, Yuan, Zhang, and Feng]{wang2019distilling}
Tao Wang, Li Yuan, Xiaopeng Zhang, and Jiashi Feng.
\newblock Distilling object detectors with fine-grained feature imitation.
\newblock In \emph{Proceedings of the IEEE/CVF Conference on Computer Vision and Pattern Recognition}, pages 4933--4942, 2019.

\bibitem[Wang et~al.(2022)Wang, Zhang, Yang, and Sun]{wang2022anchor}
Yingming Wang, Xiangyu Zhang, Tong Yang, and Jian Sun.
\newblock Anchor detr: Query design for transformer-based detector.
\newblock In \emph{Proceedings of the AAAI Conference on Artificial Intelligence}, pages 2567--2575, 2022.

\bibitem[Wang et~al.(2024)Wang, Li, Weng, Zhang, Yue, Feng, Han, and Ding]{wang2024knowledge}
Yu Wang, Xin Li, Shengzhao Weng, Gang Zhang, Haixiao Yue, Haocheng Feng, Junyu Han, and Errui Ding.
\newblock Kd-detr: Knowledge distillation for detection transformer with consistent distillation points sampling.
\newblock In \emph{Proceedings of the IEEE/CVF Conference on Computer Vision and Pattern Recognition}, pages 16016--16025, 2024.

\bibitem[Yang et~al.(2022{\natexlab{a}})Yang, Li, Jiang, Gong, Yuan, Zhao, and Yuan]{yang2022focal}
Zhendong Yang, Zhe Li, Xiaohu Jiang, Yuan Gong, Zehuan Yuan, Danpei Zhao, and Chun Yuan.
\newblock Focal and global knowledge distillation for detectors.
\newblock In \emph{Proceedings of the IEEE/CVF Conference on Computer Vision and Pattern Recognition}, pages 4643--4652, 2022{\natexlab{a}}.

\bibitem[Yang et~al.(2022{\natexlab{b}})Yang, Li, Shao, Shi, Yuan, and Yuan]{yang2022masked}
Zhendong Yang, Zhe Li, Mingqi Shao, Dachuan Shi, Zehuan Yuan, and Chun Yuan.
\newblock Masked generative distillation.
\newblock In \emph{Proceedings of the European Conference on Computer Vision}, pages 53--69. Springer, 2022{\natexlab{b}}.

\bibitem[Zhang et~al.(2023)Zhang, Li, Liu, Zhang, Su, Zhu, Ni, and Shum]{zhang2023dino}
Hao Zhang, Feng Li, Shilong Liu, Lei Zhang, Hang Su, Jun Zhu, Lionel~M Ni, and Heung-Yeung Shum.
\newblock Dino: Detr with improved denoising anchor boxes for end-to-end object detection.
\newblock In \emph{Proceedings of the International Conference on Learning Representations}, 2023.

\bibitem[Zhang and Ma(2020)]{zhang2020improve}
Linfeng Zhang and Kaisheng Ma.
\newblock Improve object detection with feature-based knowledge distillation: Towards accurate and efficient detectors.
\newblock In \emph{Proceedings of the International Conference on Learning Representations}, 2020.

\bibitem[Zhang et~al.(2020)Zhang, Chi, Yao, Lei, and Li]{zhang2020bridging}
Shifeng Zhang, Cheng Chi, Yongqiang Yao, Zhen Lei, and Stan~Z Li.
\newblock Bridging the gap between anchor-based and anchor-free detection via adaptive training sample selection.
\newblock In \emph{Proceedings of the IEEE/CVF Conference on Computer Vision and Pattern Recognition}, pages 9759--9768, 2020.

\bibitem[Zheng et~al.(2022)Zheng, Ye, Wang, Ren, Zuo, Hou, and Cheng]{zheng2022localization}
Zhaohui Zheng, Rongguang Ye, Ping Wang, Dongwei Ren, Wangmeng Zuo, Qibin Hou, and Ming-Ming Cheng.
\newblock Localization distillation for dense object detection.
\newblock In \emph{Proceedings of the IEEE/CVF Conference on Computer Vision and Pattern Recognition}, pages 9407--9416, 2022.

\bibitem[Zhu et~al.(2020)Zhu, Su, Lu, Li, Wang, and Dai]{zhu2020deformable}
Xizhou Zhu, Weijie Su, Lewei Lu, Bin Li, Xiaogang Wang, and Jifeng Dai.
\newblock Deformable detr: Deformable transformers for end-to-end object detection.
\newblock \emph{arXiv preprint arXiv:2010.04159}, 2020.

\end{thebibliography}
}

% WARNING: do not forget to delete the supplementary pages from your submission 
% \input{sec/X_suppl}

\end{document}